\documentclass{article}[11pt]
\usepackage{times}
\usepackage{soul}
\usepackage{url}
\usepackage[hidelinks]{hyperref}
\usepackage[utf8]{inputenc}
\usepackage[small]{caption}
\usepackage{graphicx}
\usepackage{amsmath}
\usepackage{booktabs}
\usepackage{algorithm}
\usepackage{algorithmic}
\usepackage{amssymb}
\title{Approaching Adaptation Guided Retrieval in Case-Based Reasoning through Inference in Undirected Graphical Models}
\author{
	Luigi Portinale
	\\
	Computer Science Institute, DiSIT, \\
	Universit\`{a} del Piemonte Orientale, Italy 
	\\
	{\tt luigi.portinale@uniupo.it}}
\date{}
\begin{document}
\maketitle

\begin{abstract}
In Case-Based Reasoning, when the similarity assumption does not hold, the retrieval of a set of cases structurally similar to the query does not guarantee to get a reusable or revisable solution.
Knowledge about the adaptability of solutions has to be exploited, in order to define a method for adaptation-guided retrieval. We propose a novel approach to address this problem, where knowledge about the adaptability of the solutions is captured inside a metric Markov Random Field (MRF). Nodes of the MRF represent cases and edges connect nodes whose solutions are close in the solution space. States of the nodes represent different adaptation levels with respect to the potential query.
Metric-based potentials enforce connected nodes to share the same state, since cases having similar solutions should have the same adaptability level with respect to the query. The main goal is to enlarge the set of potentially adaptable cases that are retrieved without significantly sacrificing the precision and accuracy of retrieval. We will report on some experiments concerning a retrieval architecture where a simple kNN retrieval (on the problem description) is followed by a further retrieval step based on MRF inference.
\end{abstract}

\section{Introduction}
\label{intro}
%
In Case-Based Reasoning (CBR), the {\em similarity assumption}
states that similar problems
have similar
solution(s).
The similarity defined on the case description is often called {\em structural similarity},
in contrast to the {\em solution similarity} defined over the solution space.
Because of that, 
the CBR problem solving process is based on the well-known
{\em 4R steps: Retrieve, Reuse, Revise and Retain} \cite{Aamodt:94}.
%
The more valid 
the similarity assumption, the more efficient the
CBR process is, since the retrieved solutions are
more similar to the (unknown) solution to the query.

The most common retrieval strategy is based on {\em k-Nearest Neighbor}
(kNN) algorithms, returning the solutions of
the $k$ cases stored in the library
that are most similar (from the structural point of view) to the query.
However, the similarity assumption is not always guaranteed to hold, and it has been
questioned several times \cite{Smyth:agr:98,Stahl02,Aziz:14}.
%
%
%
Adaptation guided retrieval can be exploited when it is not possible to rely only on structural similarity.
Different solutions have been devised in this context: the introduction of specific or task dependent adaptation knowledge into the
retrieval step \cite{Smyth:agr:98,Portinale:cbr:97,Diaz:03}, the modeling of solution preferences in preference-based CBR \cite{Aziz:14},
the learning of a utility-oriented similarity measure minimizing the discrepancies between the similarity values and the desired utility scores 
\cite{Xiong:06}.

In this paper, we describe and experiment with a novel technique where standard kNN retrieval is complemented through an adaptation guided inference process; in particular,
we propose 
to model adaptation knowledge through the construction of a
particular type of undirected graphical model, a metric
Markov Random Field (MRF) \cite{Murphy:ugm}, and
to exploit MRF inference to enhance the
retrieval step in terms of more adaptable cases.
%

The paper is organized as follows: section~\ref{mrf} outlines some basic
notions concerning MRFs and metric MRFs; 
section~\ref{adap} describes the characterization of case adaptability
we rely on; section~\ref{framework} discusses a framework for case
retrieval based on inference on an MRF capturing the relevant
adaptation knowledge for the cases of interest; section~\ref{architecture}
proposes a possible retrieval architecture integrating kNN retrieval
with the MRF inference enhancement step; section~\ref{experiments}
introduce the experimental framework for the evaluation of the proposed
architecture, section~\ref{results} reports the results 
which are finally discussed in section~\ref{conclusions}.

\section{Markov Random Fields}
\label{mrf}
A {\em Markov Random Field} (MRF) is a undirected graphical model defined as the pair $\langle \cal{G}, \cal{P} \rangle$ where
$\cal{G}$ is an undirected graph whose nodes represent random
variables (we assume here discrete random variables) and edges represent dependency relations among connected variables (i.e. an edge between
$X_i$ and $X_j$ means that $X_i$ and $X_j$ are dependent variables);
$\cal{P}$ is a probabilistic distribution over the variables represented in 
$\cal{G}$.
We restrict our attention to {\em pairwise MRFs}:
each edge $(X_i$ --- $X_j)$ is associated with a {\em potential}
$\Phi_{i,j}: D(X_i) \times D(X_j) \rightarrow \mathbb{R}^+ \cup \{0\}$;
here $D(X)$ is the domain (i.e. the set of states or values) of the
variable $X$.

In a MRF, the distribution $\cal{P}$ factorizes over $\cal{G}$; this means
that
\[\mathcal{P}(X_1 \ldots X_n)=\frac{1}{Z} \prod_{i,j} \Phi_{i,j}(X_i, X_j) \]
where $Z=\sum_{X_1 \ldots X_n} \prod_{i,j} \Phi_{i,j}(X_i, X_j)$ is a
normalization constant called the {\em partition function}.

We consider a special case of pairwise MRF called
{\em metric MRF}. In a metric MRF, all nodes take values in the
same label space $V$, and a distance function $d: V \times V \rightarrow \mathbb{R}^+ \cup \{0\}$ is defined over $V$;
the edge potentials are defined as
\[\Phi_{i,j}(x_i, x_j)=\exp(-w_{ij} d(x_i,x_j))\]
given that $x_i$ is a value or state of variable $X_i$, and
$w_{ij}>0$ is a suitable weight stressing the importance of the
distance function in determining the potential.
The idea is that adjacent variables are more likely to
have values that are close in distance.

Concerning inference, 
we are interested in the computation
of the posterior probability distribution
of each single unobserved variable, given the evidence.
In this paper, we will resort to {\em mean
field inference}, a variational approach where the target distribution is
approximated by a completely factorized distribution \cite{Weiss:00}.
This algorithm is implemented in the {\sc Matlab} UGM toolbox \cite{UGM}
that we have exploited in our experimental analysis.

\section{Case Solutions and Adaptation Knowledge}
\label{adap}

The standard CBR process requires the definition
of a suitable distance metric over the case description (i.e. the
case features) and it employs a kNN algorithm, in order to get
the k most structurally similar cases. Given the
similarity assumption, we expect that the solutions of similar cases are also similar. If this assumption is not valid, kNN retrieval can result
is a set of useless cases, since their adaptation level with respect
to the query is unsuitable (i.e., no adaptation mechanism can be either
devised or adopted with a reasonable effort).
If this is the case, some abstract notion of adaptation space should be
defined, and
adaptation knowledge has to be exploited
during retrieval (see \cite{Leake:97,Bergmann:16}).
A common approach is to consider
the solution space equipped with a similarity or a distance metric
\cite{Stahl02,Aziz:14}.
This metric can then be used to measure how close two potential
solutions are: the closer the solutions, the more similar the adaptation effort needed to revise them for a specific query.
When adapting a solution with respect to a query, 
we need to use the available adaptation knowledge; depending on the
``complexity'' of the inference process executed during adaptation, we can
devise different adaptation levels corresponding to the effort (or cost) needed
to perform this phase. For example, if no adaptation is needed (i.e., the query case is solved through the reuse step) we can map this situation to the
minimum level of adaptation effort. On the contrary, suppose that the adaptation knowledge is provided through adaptation rules: the type and the number of applied rules can define a measure of the effort needed to revise the solution, 
and such a measure can be mapped into different adaptation levels (see below).

\medskip

{\bf Example 1.} Suppose we have a case library containing the description
of some holiday packages
(this corresponds to the
case study {\tt Travel} described in section~\ref{experiments}); 
a customer requires a package by specifying some features as reported in Table~\ref{tab_features}, and the returned solution consists in a specific
package characterized by such features, and
completed with the name of a hotel and the package final price.
Let each hotel be described by an object $h$ with
properties $h.\mbox{Name}$ (the hotel name),  $h.\mbox{Location}$ (the hotel place) and $h.\mbox{Category}$ (the hotel classification in number of stars).
\begin{table}
\begin{center}
\begin{tabular}{|l|l|}
\hline
{\bf Name} & {\bf Domain}\\
\hline
\hline
Duration & Numeric \\
\hline
\#Persons & Numeric \\
\hline
Accommodation & Ordinal \\
\hline
Season & Ordinal \\
\hline
HolidayType & Categorical \\
\hline
Destination & Categorical \\
\hline
Transport & Categorical \\
\hline
\end{tabular}
\end{center}
\caption{\label{tab_features} Features and domains of the {\tt Travel} case base}
\end{table}
Consider now a customer with a given budget $b$ and with some specific
criteria $A(h)$ for accepting a proposed hotel $h$: if $A(h)={\tt true}$ 
then the hotel is accepted by the customer, otherwise it is rejected.
\begin{figure}
\rule{0.5\columnwidth}{0.4pt}
\begin{algorithmic}
\STATE $\mbox{{\bf Init:} } a \gets r$ ({\em retrieved case as basis for the adaptation})
\STATE $\mbox{Price\_per\_person} \gets r.\mbox{Price} / r.\mbox{\#Persons}$
({\em determine the cost for each partecipant})

\medskip
{\bf (R1)} ({\em a package can be used for less people})
\IF {$(r.\mbox{\#Persons} >q.\mbox{\#Persons})$}
	\STATE $a.\mbox{\#Persons} \gets q.\mbox{\#Persons}$
	\ENDIF

\medskip
\textbf{(R2)} (\textit{substitute coach with train with a $10\%$ increase of the price per person})
\IF {$r.\mbox{Transport=train} \wedge q.\mbox{Trasport=coach}$}
	\STATE $a.\mbox{Trasport} \gets \mbox{train}$
	\STATE $\mbox{Price\_per\_person} \gets \mbox{Price\_per\_person} * 1.1$
	\ENDIF
	
\medskip
\textbf{(R3)} (\textit{substitute train with coach with a $10\%$ decrease of the price per person})
\IF {$r.\mbox{Transport=coach} \wedge q.\mbox{Trasport=train}$}
	\STATE $a.\mbox{Trasport} \gets \mbox{coach}$
	\STATE $\mbox{Price\_per\_person} \gets \mbox{Price\_per\_person} / 1.1$
	\ENDIF 

\medskip
$a$.Price=Price\_per\_person * $a$.\#Persons

\medskip
(\textit{if price is over budget adaptation fails})
\IF {$a.\mbox{Price} > b$}
	\STATE $\mbox{flag } r \mbox{ as not adaptable and \textbf{exit}}$
	\ENDIF 
	
\medskip
\textbf{(R4)} (\textit{if hotel is not accepted, find an alternative hotel of the same category in the same place})
\IF {
$\neg A(r.\mbox{Hotel}) \wedge
\exists h (h.\mbox{Category}=q.\mbox{Accommodation}
\wedge 
h.\mbox{Location}=q.\mbox{Destination} \wedge
h.\mbox{Name} \neq r.\mbox{Hotel})$}
	\STATE $a.\mbox{Accommodation} \gets h.\mbox{Category}$
	\STATE $a.\mbox{Destination} \gets h.\mbox{Location}$
\ELSE 
	\STATE $\mbox{flag } r \mbox{ as not adaptable}$
\ENDIF
\end{algorithmic}
\rule{0.5\columnwidth}{0.4pt}
\caption{\label{adap_rules}Adaptation rules for the case study.}
\end{figure}
Given a query (with the specification of some of the features in 
Table~\ref{tab_features}), suppose we have the adaptation rules reported
in Figure~\ref{adap_rules} (where $r$ refers to the retrieved case, $q$ to the query, $a$ to the adapted case and $h$
to a particular hotel)
to be applied in sequence.
Given the above adaptation rules,
we can in principle define different adaptation levels by considering the type
and the number of rules applied during revise. By way of example,
we could devise the following levels:
\begin{itemize}
\item
Level 1: if no rule is applied
\item
Level 2: if only rule R1 is applied
\item
Level 3: if one or more rules R2, R3, R4 are applied with success
\item
Level 4: if case is flagged as not adaptable
\end{itemize}
For instance, we can consider a solution with level 1 to be reusable,
a solution with level 2 to be revisable with a small cost,
a solution with level 3 revisable with a larger cost and a solution with level 4 to be
unadaptable.
%
In the following, we propose to characterize the principle
``similar solutions imply similar adaptation levels'' by means of a metric
MRF built on a given case library; 
section~\ref{framework} will discuss how MRF inference can be then
exploited to provide an adaptation guided retrieval approach.

\section{MRF Inference for Retrieval of Adaptable Cases}
\label{framework}

Given a case library of stored cases with solutions, we 
first construct a metric MRF.
Let 
{\tt \#adapt\_levels}
be the number of different adaptation levels, $sim\_s$ be the
similarity metric defined over the solution space and $st$ be a threshold
of minimal similarity for solutions.
Algorithm~\ref{mrf_alg} shows the pseudo-code for the construction
of the MRF.

\begin{algorithm}[t]
\caption{\label{mrf_alg}MRF construction.}
\begin{algorithmic} 
    \REQUIRE ${\tt \#adapt\_levels}; st > 0$
    \ENSURE a metric MRF
    \STATE $\mbox{MRF} \gets \mbox{empty graph}$
    \FOR{$\mbox{each case } c$}
    		\STATE add node $c \mbox { to MRF}$
    \ENDFOR
    \FOR{$\mbox{each node } n$}
    		\STATE $num\_states(n) \gets {\tt \#adapt\_levels}$
	\ENDFOR
    \FOR  {$\mbox{each pair of nodes } (n,m)$}
    		\IF {$sim\_s(n,m)>st$} 
    			\STATE $\mbox{add edge } e=(n,m) \mbox{ to MRF}$
    		\ENDIF
	\ENDFOR
	\FOR{$\mbox{each edge } e=(n,m)$}
		\STATE $s \gets sim\_s(n,m)$
		\FOR{$i=1 \ldots {\tt \#adapt\_levels}$}
			\FOR{$j=1 \ldots {\tt \#adapt\_levels}$}
				\STATE $\Phi_{n,m}(i,j) \gets \exp(-s \; |i-j|)$
			\ENDFOR
		\ENDFOR
	
	\ENDFOR
    
\end{algorithmic}
\end{algorithm}
The idea is to build a metric MRF where nodes have the possible case
adaptation levels as states 
and they are
connected only if the corresponding cases have a sufficiently large solution similarity
(greater than the threshold $st$).
Edge potentials $\Phi_{n,m}(i,j)$
are determined as in metric MRFs: the smaller the distance between
the states $i$ and $j$ of nodes $n$ and $m$ respectively, the larger
the related potential entry. 
When a given node assumes a specific state, connected nodes
tend to assume close state values with a high probability (i.e.,
 if a case has a given adaptation level, we expect the
cases having similar solution to have a very
close adaptation level).
%
Moreover, the more similar the solutions of the cases, the stronger this effect should be; 
this is the reason why we use solution similarity $s=sim\_s(n,m)$ as a
weight for the metric potential.
Once we know the adaptation level of some of the stored cases,
MRF inference can be used to propagate this information in the
case library; this for identifying other cases as good candidates for solution reuse or revision, as well as rejecting some cases because they are likely to be
useless for adaptation or reuse.
The idea is then to start from a standard kNN retrieval, followed
by the usual reuse and revise steps. 
The results
of the reuse/revise phases are used as input for MRF inference.
%
Algorithm~\ref{mrf_inf} details this process.
\begin{algorithm}
\caption{\label{mrf_inf}MRF inference for adaptable cases retrieval.}
\begin{algorithmic} 
    \REQUIRE $al(1) \ldots al(k); cond()$
    \ENSURE a set of (cases, adaptation levels) pairs $(c, al)$
    \FOR{$i=1:k$}
    		\STATE $set\_evidence(i,al(i))$
     \ENDFOR 
     \STATE $Bel[\;]=MRF\_Inference$
     \FOR{each not retrieved case $c$}
    		\IF {$cond(Bel[i])$}
    			\STATE output $(i, cond(Bel[i]))$
    		\ENDIF
     \ENDFOR

\end{algorithmic}
\end{algorithm}
Let $al(i)$ be the adaptation level of the $i$-th retrieved case (through
kNN retrieval);
for each retrieved case, its adaptation level is set as evidence in the
corresponding node of the MRF. Inference is then performed and the posterior probability of each MRF node is computed into the multidimensional vector
$Bel$. $Bel[i]$ is the posterior distribution or {\em node belief}
of node $i$; $Bel[i]$ is a $l$-dimensional vector (where $l$ is
the number of adaptation levels) such that $Bel[i,j]$ is
the probability of node $i$ being in state $j$ given the evidence.
Input parameter $cond()$ is a function
testing a condition on the node belief; if this condition is not satisfied, it returns {\tt false}, otherwise it returns the state of the node (i.e., the
adaptation level) for which the condition is satisfied.
Algorithm~\ref{mrf_inf} finally outputs a set of cases with their adaptability
level.
\medskip

{\bf Example 2.} Suppose we want to determine for each (non retrieved)
case the most probable adaptation level, then we will set
$cond(Bel[i])=\arg\max_j Bel[i,j]$.
In this case every case has a potential adaptation level and we
can consider it for the next actions: for instance, we could
be interested only in the most easily adaptable cases, and if
$1$ is the minimum adaptation level, we
will select only those nodes $i$ for which $cond(Bel[i])=1$

Consider now a more complex condition: suppose we consider as
interesting
any adaptation level from $1$ to $a$, and suppose that we want
to be pretty sure about the adaptability of the case. We could set
a probability threshold $pt$ and to require that


\begin{algorithmic}
\IF{$Bel[i,1]+ \ldots Bel[i,a]>pt$}
	\STATE $cond(Bel[i]) \gets a$
\ELSE 
	\STATE $cond(Bel[i]) \gets {\tt false}$
\ENDIF
\end{algorithmic}

In this case we are collapsing all the adaptability levels from 1 to $a$
into a unique level (the choice of $a$ is completely arbitrary here, and any
other label would be fine as
we no longer need to distinguish them); in case the required confidence
on adaptability is not reached, we will simply ignore the case.
Of course, several other implementations of the $cond()$ function can be devised.

\section{An Architecture for Adaptation Guided Retrieval}
\label{architecture}


The problem of retrieving ``useful'' cases with respect to
a given query is characterized by
two different aspects: the structural similarity between the query and
the retrieved case (addressed by kNN retrieval), and
the adaptability to the query of the retrieved solution (addressed
by MRF inference); this means that the cases of interest
are those which are sufficiently similar to the query, while having an adaptable solution. We call them {\em positive cases}.
We would like the retrieval to return only positive cases, possibly
with a large structural similarity and with a low adaptability cost (i.e., a low
adaptability level).
While cases retrieved through kNN do not have the guarantee of
being adaptable, cases retrieved through MRF inference are
more likely to be adaptable, but they do not
have any guarantee of being sufficiently similar to the query.
Moreover, the reason why retrieval is often restricted to a set of $k$
cases, is because it is in general unfeasible to take into
consideration all the positive cases: considering all the cases
returned by MRF inference may lead to an unreasonably
large number of cases to be managed.

The proposed retrieval architecture starts with standard kNN retrieval;
if  all the $k$ retrieved cases are actually adaptable, then the process
terminates with such $k$ cases as a result. 
On the contrary, let $0 \leq k' < k$ be the number of adaptable
cases retrieved by kNN. Since there is still room for
finding positive cases, MRF inference 
is performed as shown in Algorithm~\ref{mrf_inf}, then the top
$k-k'$ cases in descending order of structural similarity are returned,
from
the output of Algorithm~\ref{mrf_inf}.
The main idea underlying this architecture is that $k$ is the
desired output size. In case kNN retrieval tangles with the solution similarity
problem, we complement the retrieval set with some cases
that are likely to be adaptable. Since they are selected by considering
their structural similarity with respect the query, we also
maximize the probability
of such cases being positive.
In order to evaluate the effectiveness of such architecture, we
set up an experimental framework described
in section~\ref{experiments}, and which results are reported in section~\ref{results}.

\section{Experimental Testbed: the {\tt Travel} dataset}
\label{experiments}

As a testbed for the approach,
we consider a dataset called {\tt Travel} containing instances
of about $1500$ holiday packages described through the features reported in
Table~\ref{tab_features}. Such features are those used to
query the case base; in addition, stored cases also contain the price of the package
and the name of a hotel. 

Local distances for features are defined as follows:
for numeric features and for the ordinal attribute Accommodation (mapped into
integers from $0$ to $5$)
we adopted the \textit{standardized Euclidean distance} (Euclidean distance normalized by feature's standard deviation);
for categorical features we adopted the \textit{overlap distance} ($0$ if two values are equal and $1$ if they are different); finally for the ordinal attribute Season
we adopted a \textit{cyclic distance}, since the values are mapped into the ordinal numbers of the months. The cyclic distance on a feature $f$ is defined by the following formula
$d_f(x,y)=\min(|x-y|, R_f-|x-y|)$ where $x,y$ are the values (from $1$ to $12$ in such a case) and $R_f=\mbox{range}(f)+1$ ($12$ in this case).
In all the above cases, when there is a missing value, the maximum distance value for the feature is considered.
We also have defined a vector of feature weights and the aggregation function producing
the global distance $d(i,j)$ between two cases $i$ and $j$ is the weighted average.

The solution of a case is actually a complete description of a package (including a specific hotel and the computed price).
Since adaptation knowledge considers the price and the hotel characteristics, in order
to revise a potential solution, the characterization
of the similarity of solutions only requires information about
the price, the destination and the the hotel category.
Distance between two solutions is again computed as a weighted average
of the three local distances on Price, Accommodation and Destination.

Similarity measure for both case descriptions and solutions is 
computed as
\[s(i,j)=\frac{1}{1+d(i,j)}\]
where $d()$ is the global distance
($0 < s(i,j) \leq 1$).


We consider the example adaptation rules illustrated in Section~\ref{adap};
we set a budget $b$ on price equal to the average price of the packages in the
case base; moreover, in order to simulate the acceptability criterion, we implemented a random acceptance with probability of $80\%$.
Concerning the adaptability levels, we consider a binary characterization with only
two levels: $1=\mbox{adaptable}$, $2=\mbox{not adaptable}$.
In our experiments, we also set the threshold $pt=0.9$  
(i.e., MRF inference considers a case as adaptable if the probability of the corresponding MRF node of assuming state $1$ is greater than $90\%$).
Finally, we define
a similarity threshold $thr$ on the structural similarity between a retrieved
case and a query. 
A ``positive'' case is a case having
a structural similarity with the query greater than or equal to $thr$, and
such that its adaptability level is $1$.
%
%
With the above characterization of positive cases, 
given a particular retrieval set, we consider the usual notions of {\em accuracy},
{\em precision}, {\em recall} (more precisely precision and recall at $k$ since
we focus on retrieving $k$ cases) and {\em F1}-{\em measure}
(the harmonic mean of precision
and recall).

\section{Experimental Results}
\label{results}
%
If in a problem there are features that are predictive of the solution,
then we expect that similar values
of such features will corresponds to similar values of the solution.
In order to stress the  ``similarity assumption'' and
to bring out problems related to it,
we have considered queries with missing values
for such ``solution-correlated'' features.
We decided to evaluate the performance of the proposed architecture
by selecting as potential missing features the attiributes Duration, Accommodation and HolidayType; taken together they have a correlation coefficient with the Price (the most relevant part of the solution) of $0.77$, thus it can be expected
that, missing values on such features will reduce the validity of the similarity assumption.

For a given query, we set the probabilities of a missing value
as follows: $p_A=0.15$ for Accommodation, $p_D=0.3$ for Duration
and $p_H=0.6$ for HolydayType.
In all the experiments we adopted the thresholds $st=0.9$
(see Algorithm~\ref{mrf_alg}) and $pt=0.9$ (see
Example~2 in section~\ref{framework}).
%
The construction of the metric MRF (Algorithm~\ref{mrf_alg}) and
the MRF inference (Algorithm~\ref{mrf_inf}) have been implemented in
{\sc Matlab} by exploiting the UGM toolbox \cite{UGM}. In particular,
we resort to mean field variational inference as previously mentioned\footnote{ Comparable results have been obtained by using
Loopy Belief Propagation \cite{Weiss:00}.}.
We finally consider two
other evaluation parameters,
the structural similarity threshold $thr$ and the number of retrieved cases $k$, and
we perform different runs by varying such parameters.
In particular,
we set $thr=\alpha \mu_c$ where $\mu_c$ is
the average structural similarities among
the cases in the case library, and
$\alpha$ is a scale factor.
Variation on $thr$ influences the set of positive cases\footnote{
Others parameters that can be used to vary the set of positive cases are those influencing
adaptation, namely the customer budget $b$ and the hotel acceptability criterion $A(h)$; for the sake of brevity we do not consider them here.} while
variation on $k$ models different retrieval capabilities.
We set values of $k$ from 1 to 15 with step of 1, and then
from 20 to 100 with step of 10.
Greater is the threshold $thr$, smaller is the set of positive cases
and vice-versa;
values of $\alpha$ are chosen in such a way of setting the threshold at $75\%$
of the average similarity, at the median of the case similarities
($95\%$ of the average in this problem), at the average similarity, and finally
setting the threshold $25\%$ and $50\%$ above the average similarity.
In particular, given a case base of $1500$ cases, the average number
of positive cases resulted as follows: $559$ for $\alpha=0.75$, 
$400$ for $\alpha=0.95$, $320$ for $\alpha=1$, $86$ for $\alpha=1.25$, $11$ for $\alpha=1.5$.
A few words are needed to explain the choice of the considered $k$ values:
small values are reasonable when the query should not return too many results to the user (as in the case of recommending holiday packages). However, our aim here is to use the {\tt Travel} dataset as an evaluation framework, not tied to the specific application task of recommending travels; in situations
where the scale factor $\alpha \leq 1$, the number of relevant cases (i.e., positive
cases) is rather large, producing a very low recall for small values of $k$. For
this reason we considered large values of the $k$ parameter (from 20 to 100) as well.

We have performed a 10-fold cross validation for every considered value of $k$,
by measuring every time the mean values for
accuracy, precision, recall and F1-measure,  for
both simple kNN retrieval (label kNN in the figures)
and kNN+MRF inference (label MRF in the figures).
\begin{figure}[htb]
\centering
\includegraphics[width=1.0\columnwidth]{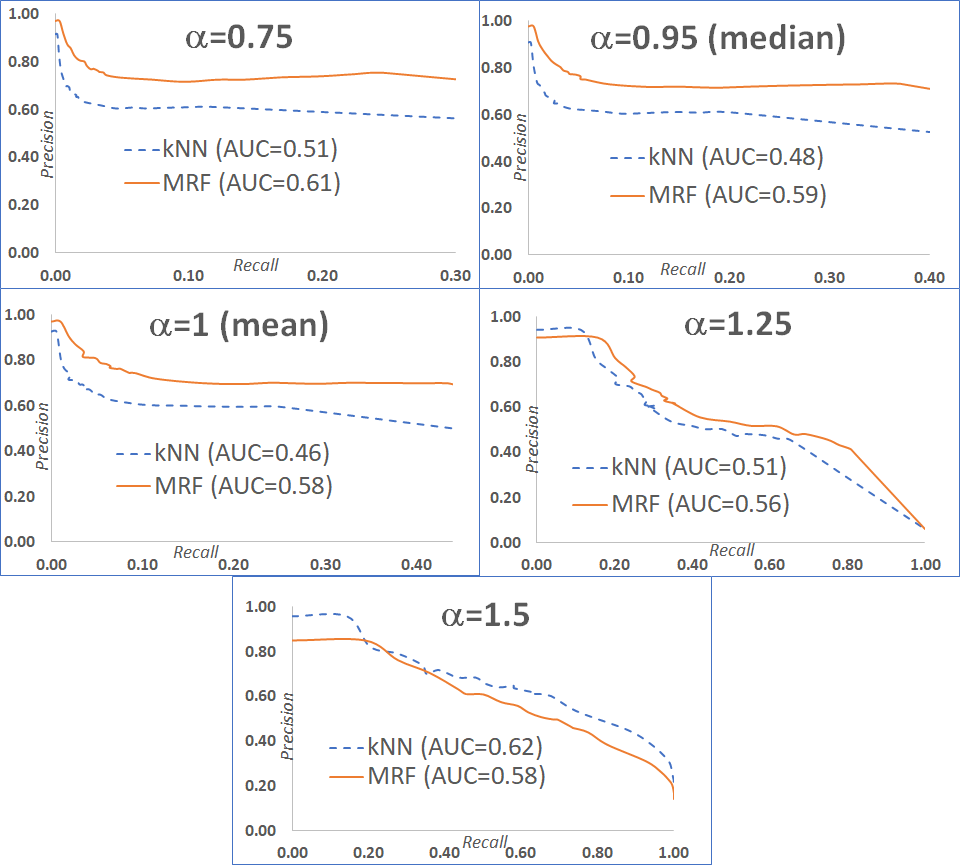}
\caption{\label{prc} PR Curves for different scale factors}
\end{figure}
Figure~\ref{prc} shows the Precision/Recall (PR) curves obtained for the different values
of the scale factor $\alpha$, by varying the value of $k$: increasing the value
of $k$ will increase recall, by decreasing precision.
The curves are actually an approximation of the real PR curves, since in the experiments we almost never reached a situation with a recall close to $1$\footnote{
To obtain this result we should have considered a very large value for parameter $k$; indeed,
when the set of positive cases is large ($\alpha \leq 1$ in our experiments), recall
is necessarily low, and can be increased only with large values of $k$.};
as usual in these
cases, we set the last point of the PR curve with the pessimistic estimate for precision corresponding to $\frac{P}{N}$,
where $P$ is the number of positive cases, and $N$ the total number of cases.
An exception to this situation is the case $\alpha=1.5$, where with $k=90$ we get an average recall very close to $1$, producing the PR curve shown in the bottom right of Figure~\ref{prc}.
We also computed the Area Under the Curve (AUC); for the reasons outlined above, the computed value is a pessimistic estimate (smaller than the actual one), but
in the case of $\alpha=1.5$ where it has been possible to compute it exactly.
The graphics of Figure~\ref{prc} with $\alpha \leq 1$ only show a part
of the PR curve, since the estimated last point $(1,\frac{P}{N})$ would result really
far from the last measured point (the reported values for AUC are however computed by taking into account the whole curve).


%
%

%
%

\begin{figure}[htb]
\centering
\includegraphics[width=1.0\columnwidth]{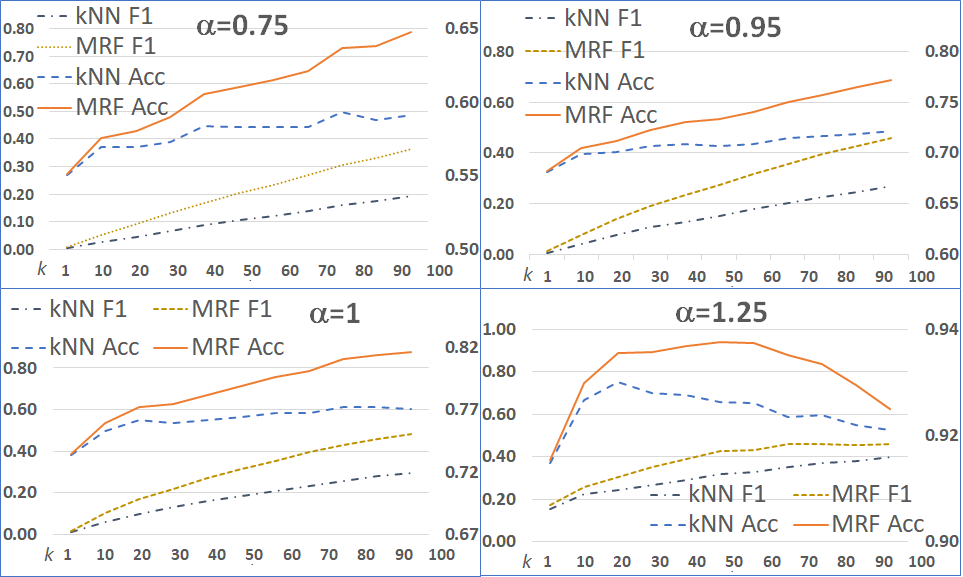}
\caption{\label{acc} Accuracy and F1-score vs $k$ for different scale factors.}
\end{figure}

Figure~\ref{acc} shows the behavior of accuracy and F1 measure, in dependence of $k$, for different values of $\alpha$.
Values for the accuracy are plotted on the right axis, values for F1 on the left axis.

\begin{figure}[htb]
\centering
\includegraphics[width=1.0\columnwidth]{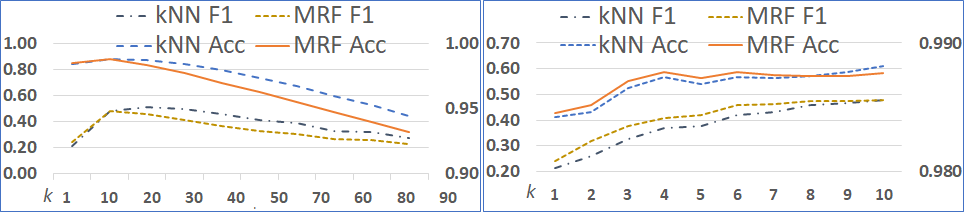}
\caption{\label{accsmall} Accuracy and F1-score for $\alpha=1.5$}
\end{figure}

%
Accuracy and F1 measure for the specific case of $\alpha=1.5$
are reported in Figure~\ref{accsmall}
(again, values for the accuracy are plotted on the right axis, values for F1 on the left one).
In particular, we report the whole plot ($k=1 \ldots 90$) on the left part of the figure,
and we ``magnify'' the plot for  $k=1 \ldots 10$ on the right part.

\section{Discussion and Conclusions}
\label{conclusions}
From the experimental results we notice that the benefits 
of the MRF approach with respect to
simple kNN retrieval (in terms of balance
between precision and recall) are all the more evident how large is
the size of the set of positive cases (i.e., for small values of $\alpha$).
This is noticeable from both PR curves (and the corresponding values for AUC), and
F1 measure.
When the are a a lot of cases sufficiently similar
to the query and potentially adaptable, kNN alone has difficulty in retrieving positive cases: simple structural similarity is not
sufficient and
the integration with MRF inference is fruitful.

The MRF integration also provide benefits in terms of accuracy as shown in
Figure~\ref{acc}, since some false negatives are actually moved into true positives
with respect to simple kNN.
In general, accuracy (for both kNN and MRF) turns out to be negatively correlated
with the size of the set of positive cases. Moreover, when 
the number of retrieved cases $k$ is too large with respect
to the number of positives, accuracy shows a decreasing pattern as we can expect,
since too many false positives can be potentially retrieved
(this is evident in the plots relative to $\alpha=1.25, 1.5$).

A better behavior of kNN is apparent in case of $\alpha=1.5$, but it is worth noting
that in this situation we have very few positive cases ($11$ on average in each run), making not very significant the results for large values of $k$.
This is the reason why in Figure~\ref{accsmall}
we considered also the situation restricted to
$1 \leq k \leq 10$; by considering reasonable values for $k$, even in the
case of $\alpha=1.5$, both accuracy and F1 measure show a small advantage
in adopting the MRF integration to kNN retrieval.
In conclusions, the evaluation in terms of accuracy, precision and recall of the considered testbed suggests that the proposed integrated architecture can provide
advantages, 
when simple structural similarity is not able to suitably capture 
the actual effort in adapting the retrieved solutions to the current query.
A final remark is worth on the characteristics of the MRF models that have been obtained in the study; they are undirected graphs that tend to have multiple connected components, since only cases having close solutions are
connected, naturally resulting in different independent groups of nodes (see Figure~\ref{mrf} for a typical example).
\begin{figure}[htb]
\centering
\includegraphics[width=1.0\columnwidth]{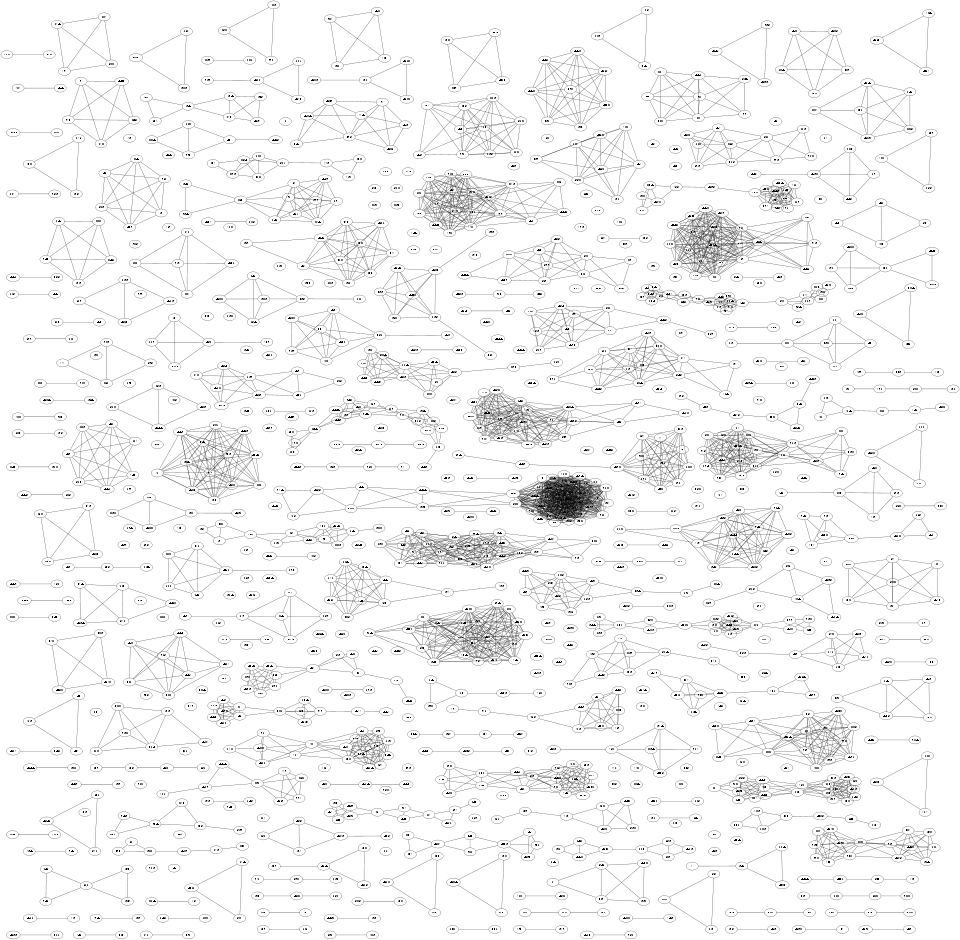}
\caption{\label{mrf} A typical MRF with node groups}
\end{figure}
This means that even if the number of cases becomes large,
the approach is likely to scale-up; inference on such groups can be performed independently, and if
they involve a limited number of nodes, even exact inference may be attempted.


The integration of CBR
and graphical models for retrieval has been usually investigated by concentrating on directed
models like Bayesian Networks (BN).
In \cite{Aamodt:bn:98},
a BN model is coupled with a semantic network
to adress case indexing and retrieval.
%
BN-based retrieval is triggered by the introduction of the observed features
as evidence, and cases can be set in a particular {\tt on} state and retrieved if the posterior probability of such a state exceeds a given threshold.
Recent advances in this setting are presented in
\cite{Aamodt:bn:18} within the BNCreek system which
applies a Bayesian analysis aimed at
increasing the accuracy of the similarity assessment. 
These approaches 
focus only on structural similarity and 
there is no attempt to address the problem of adaptation-guided retrieval.
Our approach can then be seen as  the first attempt of exploiting
probabilistic inference on a graphical model to build a strategy
for adaptation guided retrieval.


\bibliographystyle{plain}
\bibliography{arxiv}

\end{document}